\UseRawInputEncoding
 \pdfoutput=1
 \pdfoutput=1
 \pdfoutput=1
 \pdfoutput=1
 \pdfoutput=1
\documentclass[runningheads]{llncs}

\usepackage{eccv}

\usepackage{eccvabbrv}
\usepackage{silence}

\usepackage{graphicx}
\usepackage{booktabs}

\usepackage[accsupp]{axessibility}  

\usepackage{hyperref}

\usepackage{float}
\usepackage{xcolor}
\usepackage[linesnumbered,ruled,vlined]{algorithm2e}
\definecolor{commentcolor}{rgb}{0.2, 0.6, 0.2} 
\newcommand{\PyComment}[1]{\ttfamily\textcolor{commentcolor}{\# #1}}
\newcommand{\PyCode}[1]{\ttfamily\textcolor{black}{#1}}
\usepackage{authblk}

\usepackage{orcidlink}

\begin{document}

\title{From Pretext to Purpose: Batch-Adaptive Self-Supervised Learning} 

\titlerunning{Batch-Adaptive Self-Supervised Learning}

\author{Jiansong Zhang\thanks{These authors contributed equally to this work}\inst{1,2}\orcidlink{0000-0002-6960-0667}\and
Linlin Shen$^*$\inst{1}\and
Peizhong Liu\thanks{Corresponding author}\inst{2}}

\authorrunning{Jiansong Zhang, Linlin Shen, et al.}

\institute{School of Computer Science \& Software Engineering, Shenzhen University, China\and
School of Medicine, Huaqiao University, China\\
\email{jasonzhang3111@gmail.com, llshen@szu.edu.cn, pzliu@hqu.edu.cn}}

\maketitle

\begin{abstract}
Recently, self-supervised contrastive learning has become a prominent paradigm in artificial intelligence. This approach enables unsupervised feature learning by contrasting instance-level data. However, developing an effective self-supervised learning paradigm remains a key challenge in this field. This paper starts from two important factors affecting self-supervised learning, namely batch size and the design of pretext tasks, and proposes an adaptive self-supervised learning method that integrates batch information. The proposed method, via dimensionality reduction and reconstruction of batch data, enables formerly isolated individual data to partake in intra-batch communication through the Embedding Layer. Moreover, it adaptively amplifies the self-supervised feature encoding capability as the training progresses. We conducted a linear classification test of this method based on the classic contrastive learning framework on ImageNet-1k. The empirical findings illustrate that our approach achieves state-of-the-art performance under equitable comparisons. Benefiting from its "plug-and-play" characteristics, we further explored other contrastive learning methods. On the ImageNet-100, compared to the original performance, the top1 has seen a maximum increase of 1.25\%. We suggest that the proposed method may contribute to the advancement of data-driven self-supervised learning research, bringing a fresh perspective to this community.
\end{abstract}

\section{Introduction}
\label{sec:intro}
Because of the limitation of data labelling, unsupervised/self-supervised learning as a pattern recognition method independent of supervised learning theory has attracted much attention in recent years' research. Specifically, self-supervised learning(SSL) is a feature representation method that does not rely on external annotation. It enables feature learning from data individuals. In the past time, there has been continuous mining of new modelling paradigms dedicated to achieving the ideal feature representation under contrastive self-supervised conditions of data. Such algorithms are usually implemented in a dual-track fashion. Their general form is as follows: in a two-track network containing two encoders, data anchors are set on one side and positive and negative samples are defined as queries on the other side, respectively. Data with similar feature representations are enabled to converge to certain clusters in space during training by decreasing the distance of positive sample queries from the anchors and increasing the distance of negative sample queries from the anchors. The ability of the SSL model to provide a desirable representation of data features is influenced by several factors\cite{garrido_duality_2022}. However, for data-driven self-supervised learning, this paper argues that two key elements determine whether the SSL model can learn data features effectively, i.e., how to design the positive and negative samples (pretext task) and how much data (batch size) to use for encoder learning. This is because the agent task directly determines the form of low-level data used for SSL learning, which is a reflection of the nature of the data. Data batches, on the other hand, are the number of data instances used for learning that are loaded into the model at one time during the training process, again a key factor used to increase the upper limit originating from the data side.

The pretext task depends on the researcher's a priori understanding of the data, which is defined before training begins. Most previous studies used simple data augmentation to define positive samples from the anchor separately from negative samples outside the non-current anchor.
To achieve feature extraction efficiently, data loaders with oversized batches have been used in most of the previous studies. In contrastive learning, large data batch loading enables more positive and negative sample data to be used for model training and outputs more variable features, thus enhancing the data representation. At the same time, however, this poses an obvious challenge: it is very difficult to train a model with such a large batch size. (4096 in SimCLR\cite{chen_simple_2020} and MoCo-v3\cite{chen_empirical_2021}). This means that researchers need countless GPUs for the parallel loading of batches.  This is generally difficult to do in common scenarios and significantly raises the threshold for reproducing effective self-supervised learning frameworks. 

For the mentioned, there have been many studies that have worked on those problems and attempted to train self-supervised models with variable pretext tasks\cite{tian_contrastive_2020,tong_emp-ssl_2023} or small batches of data\cite{he_momentum_2020,chen_improved_2020,chen_exploring_2020}. However, to the best of our knowledge, there are no methods that incorporate both analyses for self-supervised comparative learning. To this end, we propose a batch fusion scheme and update negative samples in an adaptive form to add more incremental low-level representations to the training process.

To be specific, we propose a batch fusion reconstruction strategy that remodels the self-supervised signals from batches in self-supervised comparison learning without significantly changing the original learning paradigm. This enables the fused data tensor to achieve communication between all data individuals in a single batch loading. In this way, the data will no longer be used for contrast learning in the form of individual instances, but all other sample information in the same batch will be taken into account in the contrast loss calculation. We use a multi-channel 1x1 convolution with a residual module to implement communication between batches of data. The parameterised agent task design will allow the self-supervised model to adaptively mine those data forms that are beneficial for feature representation. We test the linear classification performance of other leading self-supervised learning methods of the proposed strategy in this paper on ImageNet-1K, and the method in this paper achieves the state-of-the-art self-supervised learning results known so far under fair comparison conditions.

Overall, this paper has three core contributions:
\begin{enumerate}
\item  In this paper, we merge the key elements affecting self-supervised comparative learning and propose a self-supervised learning method based on batch fusion and reconstruction. It allows inter-batch data communication to occur thereby enhancing the feature representation capability of the self-supervised learning model.
\item In this paper, we propose an adaptive module for batch communication. It receives the contrast learning loss gradient and parametrically and iteratively learns those data samples encoded in a form that is conducive to reducing the contrast learning loss to adaptively achieve agent task loading.
\item The proposed method can be used as a plug-and-play boosting method that can enhance existing self-supervised learning models and improve their representation of features without significantly increasing the number of parameters.
\end{enumerate}

\begin{figure}
  \centering
  \includegraphics[scale=1]{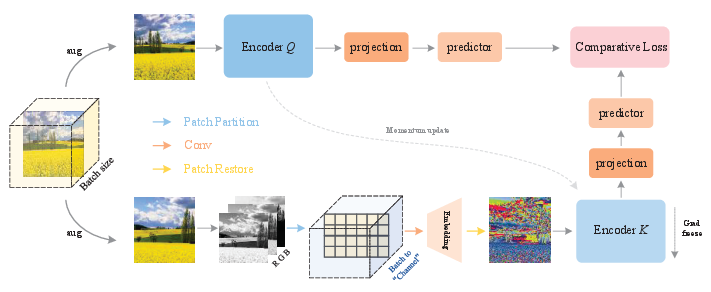}
  \caption{Overview of the proposed methods in this paper. Negative samples will be converted from 3-channel to matrix by \textbf{Pacth Partion (blue clipped head)}. Then the batch is loaded in channel form by fusion consideration into the \textbf{Embedding (orange clipped head)} shown. The output low-level matrix will be remapped to a 3-channel image via \textbf{Patch Restore (yellow clipping head)} before loading into the encoder on the negative sample side.}
  \label{figure1}
\end{figure}

\section{Related Work}
\label{sec:related}
Self-supervised learning has attracted much attention in recent years as an important feature extraction paradigm detached from labelled attributes. In this paper, we will report on the existing design paradigms for self-supervised learning from two aspects respectively.
\subsubsection{Self-Supervised Encoder Framework}In both supervised and self-supervised learning paradigms, the encoder's primary objective is feature extraction, a goal that fundamentally remains consistent across these domains. However, in the realm of label-free self-supervised feature representation learning, the design of an effective encoding framework has been a long-standing focal point of research in this field. This encompasses various works including contrastive learning\cite{chen_big_2020,chen_improved_2020,chen_exploring_2020,chen_empirical_2021,caron_emerging_2021,oquab_dinov2_2023,wu_extreme_2023}, masked autoencoders\cite{zhou_mimco_2022,xie_simmim_2022,wei_masked_2022,chen_context_2022}, and the design of loss functions\cite{ermolov_whitening_2021,zbontar_barlow_2021,tian_what_2020,ozsoy_self-supervised_2022,bardes_vicreg_2022}. Among these, the contrastive-based self-supervised learning paradigm is one of the most classic strategies\cite{garrido_duality_2022}, primarily because it allows for the acquisition of feature representations through the design of proxy tasks within a dual-track contrastive framework, thus often serving as a baseline method for testing. In recent years, the rise of masked pretext tasks has provided new insights into label-free data-driven self-supervised feature mining. The authors of \cite{he_masked_2021} and \cite{bao_beit_2022} innovatively adapted the NLP data masking pre-training approach to the domain of image self-supervised learning, reconstructing it from perspectives of image masking and positional encoding. Following this, Jinghao Zhou et.al\cite{zhou_ibot_2022} further abstracted feature representations in image self-supervised learning using a knowledge distillation-based masking learning strategy, also demonstrating the effectiveness of masking strategies in dual-track self-supervised frameworks like contrastive learning. Concurrently, the realm of non-masking pretext tasks\cite{mo_multi-level_2023,huang_revisiting_2022,oinar_expectation-maximization_2023} in self-supervised learning has witnessed numerous novel contributions. Notably, Tong et.al\cite{tong_emp-ssl_2023} employed an extremely high number of patches as a self-supervised signal, proposing a self-supervised learning framework requiring only one epoch. The remarkable success of these works is largely attributable to researchers’ deepening understanding of data processing methods in self-supervised learning.

\subsubsection{Batch in Self-Supervised Learning} Batch size is a crucial factor affecting the performance of self-supervised learning. SimCLR\cite{chen_simple_2020} was among the first to discuss the impact of batch size on self-supervised learning, noting that employing large batches directly has been regarded as a trick to enhance the performance of self-supervised frameworks. He et.al\cite{he_momentum_2020} approached the challenge from the perspective of momentum updates and queries, aiming to capture rich self-supervised signals even with smaller batches, thus facilitating effective self-supervised learning. This strategy significantly reduced the dependence on large training data batches, enabling the application of self-supervised learning in resource-constrained environments. Following this, SimSiam\cite{chen_exploring_2020} further validated the effectiveness of small batches in single-instance online contrastive learning frameworks, integrating the training benefits of BYOL\cite{grill_bootstrap_2020}. More recently, the authors of \cite{nguyen_dimcl_2023} proposed a channel-wise contrastive learning approach, endeavouring to obtain consistent and inconsistent feature representations by leveraging channel information in self-supervised training instances. Our work relates to these studies, but we focus more on efficiently utilizing batch information in self-supervised learning and integrating it into gradient computation.

\section{Method}
\label{section3}
In this section, we first describe how batch size has affected model performance in previous self-supervised learning. Immediately after that, we introduce an adaptive agent task design method for fusing batch information proposed in this paper and show that the self-supervised learning model after performing batch fusion can also serve as a powerful feature learner without the need to carefully design the agent task.
\subsection{Preliminary: Batch-based SSL}
Batch size has often been tasked in the past as an important factor that significantly affects the performance of self-supervised contrast learning. This is because, in general, contrastive learning designs, comparisons are usually performed using dual tracks and incorporating specific techniques for preventing model collapse, such as momentum updating key side encoder parameters, and freezing the gradient. However, batch size was previously considered to be a hyperparameter that could not be easily changed, regardless of the training technique added. The reason for this is that batch size significantly affects the strength of the "supervised signal" in self-supervised learning. It determines how many instances of influence the loss function needs to take into account in each iteration, thus affecting the optimisation process and ultimately the feature representation. The objective of contrastive learning training is to ensure that the encodings of positive samples, generated through model iteration, are drawn closer to the target instances, while the encodings of negative samples are pushed further away. In the context of the canonical dual-pathway contrastive learning framework, a single batch training session encompasses $N$ instances, each instance $i$ with a corresponding positive sample (a logically similar instance) and a negative sample (a logically dissimilar instance). The loss function for contrastive learning can be succinctly abstracted as depicted in Eq \ref{eq1}.
\begin{equation}
\label{eq1}
\mathcal{O}(f) = \sum_{i} \Phi \left( d(f(x_i), f(x_{i+})), \{d(f(x_i), f(x_{j}))\}_{j \neq i} \right)
\end{equation}

In the above formulas, \( f \) is the output of a parametrically encoded mapping function, \( d \) denotes a distance measure in feature space, and \( \Phi \) is a function comparing positive and negative pairwise similarities in a form that depends on the specific implementation, such as the common InfoNCE loss\cite{oord2019representation}. As the number of \(i\) increases, the negative samples involved in the computation will provide more contribution to the self-supervised loss. This means that more data individuals are comparatively taken into account in the optimisation process and increases the learning capacity of the model. However, in practice, it is often infeasible to endlessly increase the batch size, mainly due to encoder and computational resource constraints. Consequently, the question of how to efficiently mine and utilize inter-batch information, under the confines of finite computational resources and batch sizes, stands as a significant and formidable challenge. Motivated by this dilemma, this paper introduces an innovative adaptive self-supervised learning approach that harnesses batch fusion, delving into the data internally to discover solutions.

\subsection{Our Method: Batch-driven SSL}
\begin{figure}[htbp]
  \centering
  \includegraphics[scale=1]{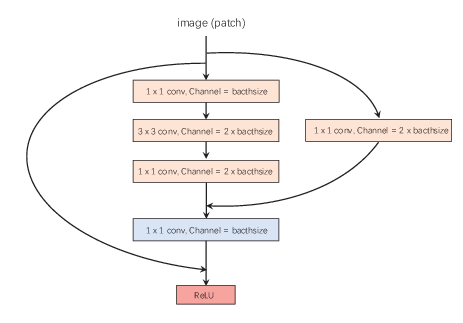}
  \caption{Specific structure of  Conv Embedding (CE).}
\label{figure3}
\end{figure}

Diverging from the widely discussed enhancements of self-supervised models in recent years, this paper primarily concentrates on data-side processing. As illustrated in Figure \ref{figure1}, it depicts the self-supervised learning architecture proposed herein. It facilitates communication among samples within a batch in a straightforward and effective manner, adaptively accomplishing self-augmentation that is beneficial for model decisions. Broadly speaking, we will elucidate how batch information is integrated across three modules to empower self-supervised learning models to perform effectively even with smaller batch sizes such as 256.
\paragraph{Patch Partiation}Taking a batch of image data with quantity $N$ as an example, where the default size of each image is $3\times224\times224$, simple augmentation implies that for a dual-path contrastive learning scenario, each path will load data of dimensions $N\times3\times224\times224$ into the self-supervised encoder. To make the batch itself noticeable in self-supervised learning, we envision employing an Embedding approach to achieve this goal. For a given image $I$, we begin by partitioning the original data into the smallest patches of size $p$. Each patch has a dimension of $3\times p\times p$, with $p$ typically set to 16. Each patch is then flattened and arranged according to its position into a two-dimensional tensor T. Specifically, for an image of size $3\times224\times224$, the dimensions will be transformed from $3\times224\times224\rightarrow196\times768$, similar to the processing in Vision Transformer (ViT)\cite{dosovitskiy2020image} as delineated in Eq \ref{eq2}.
\begin{equation}
\label{eq2}
T=\mathit{reshape}(I,[196,768])
\end{equation}
Consequently, the channel information of the image is integrated into the flattened two-dimensional tensor, while the previous batch can be regarded as "new channels". With minor adjustments, the batch then participates as a channel in the learning and decision-making processes under the self-supervised paradigm.
\paragraph{Conv and Filter}The transformation of the data tensor from three dimensions to a matrix after the reorganisation suggests that it is possible to use convolutional means to process this part. In this paper, we designed a convolutional structure, Conv Embedding (CE), as shown in Figure \ref{figure3}, which effectively enhances the communication between "channels", and contains multiple 1x1 convolutions to achieve channel expansion or compaction of the 2D tensor $T$, as shown in Eq \ref{eq2}. The \( K \)  in CE contains several 1$\times$1 convolutions to achieve channel expansion or compaction. 
\begin{equation}
\label{eq2}
T'_{c',h,w} = \sum_{c=1}^{C} K_{c',c} \cdot T_{c,h,w} + b_{c'}
\end{equation}
Through the combined operations of multiple $1\times1$ convolutions and residual connections, communication occurs between "channels," meaning the original batch information is exchanged and fused across the "channel" dimensions. Individual instances will thus access the subsequent self-supervised encoding process with rich representations from a global batch perspective. This design allows the model to retain spatial structure while enhancing feature representation through interactions between "channels". The repeated applications of $1\times1$ convolutions and residual connections not only ensure the flow of information but also provide the self-supervised encoding model with ample perspectives to learn features at different levels and degrees of abstraction.

\paragraph{Patch Restore}
In the aforementioned batch fusion structure, individual instances interact with other instances within the batch. This interaction enables the model to share and transfer information between different instances. However, to maintain the independence and integrity of each instance, a restoration structure is still necessary. It should be able to remap the two-dimensional tensor $T^{'}$, processed through multiple $1\times1$ convolutions and residual connections, back to the original image space dimensions. We refer to this process as 'Patch Restore'. Specifically, we use the inverse operation of Patch Partiation to reshape  $T^{'}$ back to the original dimensions of $3\times224\times224$. Such a procedure is similar to the mask reconstruction implemented in MAE\cite{he_masked_2021}. All patches flattened by the Patch Partiation operation are rebuilt back to the three channels in their original position. In this way, the information of each patch is restored to its original spatial position while retaining the feature information learned during the batch fusion process. Mathematically, this process can be represented as follows:
\begin{equation}
\label{eq4}
I^{'}=\mathit{reshape}({T^{'}, [3,224,224]}),\ 
I^{''}=\mathit{ReLU}(I^{'}+I)
\end{equation}
During this restoration process, we further introduce residual connections and ReLU activation functions to enhance the nonlinearity representation of the output data and optimize the quality of the reconstructed image simultaneously. Residual connections allow for a more extensive retracing of the original image's spatial domain information, while ReLU enhances those data representations that are beneficial to self-supervised learning during the adaptive process, thereby improving the model's ability to depict different features. 

Patch Restore and the enhancement of nonlinearity ensure that the size of each individual instance remains unchanged before and after batch fusion. Moreover, the data fed back to the encoder after restoration has gained more beneficial representations for self-supervised learning through the adaptive batch fusion process. This provides a robust and enriched data preparation for contrastive learning within the encoder. Our extensive experiments demonstrate that this approach significantly contributes to enhancing the performance and effectiveness of self-supervised learning.
\paragraph{Contrastive loss}Contrastive loss aims to reduce the distance between positive samples while increasing the distance between negative samples in hyperspace, thereby forming a meaningful clustering structure in the feature space. In this study, we employ the InfoNCE Loss\cite{oord2019representation} as the contrastive loss function to drive the model to learn to distinguish between positive and negative samples. It can be expressed as Eq \ref{eq6}. In this model, \( N \) denotes the batch size and \( K \) the number of negative samples. The similarity measure, denoted as \textit{sim}, can be any standard similarity function; in this paper, we default to using the $cosine$ similarity with $L2$ normalization. \( F \) represents the feature encoding function, with \( x_i \), \( x_{i+} \), and \( x_{i-} \) respectively being the instance, its positive sample, and its negative sample. The temperature \( \tau \) serves as a hyperparameter. Practically, we enhance the robustness of the loss by swapping the encoded outputs between the query and key sides, yielding a combined contrastive loss as shown in Algorithm \ref{algorithm1}.

\begin{equation}
\label{eq6}
\small
\begin{aligned}
L_{NCE} = -\frac{1}{N} \sum_{i=1}^N \log \Bigg( & \frac{\exp \left(\mathit{sim}\left(f\left(x_i\right), f\left(x_{i+}\right)\right) / \tau\right)}{\exp \left(\mathit{sim}\left(f\left(x_i\right), f\left(x_{i+}\right)\right) / \tau\right) +} \\
& \sum_{j=1}^K \exp \left(\mathit{sim}\left(f\left(x_i\right), f\left(x_{j_{-}}\right)\right) / \tau\right) \Bigg)
\end{aligned}
\end{equation}

By optimizing the InfoNCE Loss, the model is trained to increase the similarity of positive pairs while decreasing that of negative pairs. This part of the loss also influences the proposed batch fusion module, as these learnable parameters receive gradients from the InfoNCE Loss, allowing for adaptive updates during the training process. This implies that by optimizing the contrastive loss, the model not only learns to differentiate between positive and negative samples but also implicitly optimizes the information representation inherent to the data. Similar to other self-supervised learning methods, we employ various common and effective data augmentations to strengthen the self-supervised signal intrinsic to the data. These include random scale cropping, horizontal flipping, and grayscaling, among other standard augmentation techniques\cite{chen_simple_2020,grill_bootstrap_2020} in self-supervised learning.

\begin{table}[ht]
\centering
\caption{Performance comparison of the self-supervised learning method containing batch adaptive fusion technique proposed in this paper with supervised learning in each benchmark dataset. $^{*}$ indicates that we re-trained using ResNet-18.}
\label{table1}
\small
\begin{tabular}{l|l|l|l|l}
\toprule
Dataset & Method & Top 1 & Batch & Epoch \\
\midrule
CIFAR-10         &   \emph{Supervised}         &    93.15$^{*}$           &    256         &      1000      \\
                       &     Ours               &      92.12         &    256                &  1000    \\
CIFAR-100        &   \emph{Supervised}          &      69.94$^{*}$         &          256          &      1000         \\
   &     Ours               &            66.53   &             256       &     1000 \\
ImageNet-100     &   \emph{Semi-50\%}           &       75.40\cite{10.1007/978-3-031-19821-2_25}        &         -        &   -          \\
   &     Ours               &       63.62        &   256                 & 300     \\
ImageNet-1k       &   \emph{Supervised}       &       69.76$^{*}$         &           256         &     300          \\
   &     Ours               &       59.41        &   256                 &   300   \\
\bottomrule
\end{tabular}
\end{table}
\section{Experiment \& Result}
In this section, we evaluate the efficacy of the method proposed in this paper on four classic visual datasets: ImageNet-1k\cite{5206848}, ImageNet-100\cite{5206848,tian_contrastive_2020}, CIFAR-10\cite{krizhevsky2009learning}, and CIFAR-100\cite{krizhevsky2009learning}. 
\subsection{Datasets and Settings}
\paragraph{Datasets}  ImageNet-1k contains 1,000 categories with over 1,280,000 images, including 1,281,167 images in the training set and 50,000 images in the validation set. It is the most commonly used benchmark dataset for model evaluation, with natural images uniformly sized at $224\times224\times3$. ImageNet-100, a subset of ImageNet-1k, is a smaller dataset comprising 100 categories with 126,000 training samples and 50,000 test samples. Both CIFAR-10 and CIFAR-100 are medium-sized datasets with $32\times32$ resolution images. CIFAR-10 includes 10 categories with 60,000 images, 5,000 for training and 1,000 for testing per category. CIFAR-100 consists of 100 categories with 500 training images and 100 testing images per category.
\begin{table}[htbp]
\centering
\caption{Comparison of other SOTA self-supervised learning methods based on ResNet-18 and the proposed strategy in this paper. Here we use MoCo-v2 as the baseline and ensure that the comparison is done at the baseline level with the same batch size.Using \textbf{bold} formatting to highlight the best result within the same dataset.}
\label{table2}
\resizebox{\linewidth}{!}{
\begin{tabular}{llccccc}
\toprule
Method&Backbone& CIFAR-10 & CIFAR-100  & ImageNet-100 & ImageNet-1k & Batchsize   \\
\midrule
DeepCluster\cite{caron_deep_2018}  & ResNet-18 & 84.3 &50.1 &51.3 & 41.1  & 256\\
 MoCo-v2\cite{chen_improved_2020}  & ResNet-18        &  91.3      &  68.3                            &   61.9           &  52.5          & 256        \\
    SimSiam\cite{chen_exploring_2020} &    ResNet-18     &   91.2            &   64.4                 &    62.5 & 33.2   &   256       \\
        SimCLR\cite{chen_simple_2020} &      ResNet-18   &          91.1     &  65.3             &    62.1 &   52.4   & 256      \\
            BYOL\cite{grill_bootstrap_2020}&     ResNet-18    &          91.9     &   \textbf{69.2}      &  \textbf{64.1}   &   53.1  &    256     \\
                 W-MSE\cite{ermolov_whitening_2021}&     ResNet-18    &         90.6  & 64.5                &  55.8    &  -  &   256       \\
                 S3OC\cite{li2022self}&     ResNet-18    &   91.0          &        65.2       &  -    &  -  &  256        \\
                 MinEnt\cite{li2023minent} & ResNet-18 & 90.8  & 66.1 & - & - &256 \\
                    Light-MoCo\cite{lin_establishing_2023}&     ResNet-18    &   -            &            -    &  -    &  57.9  &  256        \\

                      Ours&       ResNet-18    &     \textbf{92.1}        & 66.5   &    63.6  &\textbf{59.4} &  256       \\

\bottomrule
\end{tabular}}
\end{table}

\paragraph{Implementation Details} To efficiently assess the effectiveness of the proposed method, both our method and the comparative methods utilise ResNet-18 as the feature encoder with a batch size limited to 256. We incorporate the batch fusion strategy proposed in section \ref{section3} within a two-track framework including negative samples to train the self-supervised model. The MLP has 4,096 neurons in the hidden layer and 256 in the output layer, with a temperature coefficient of 0.2 and a momentum parameter of 0.99. The training learning rate is set at 1.5E-4, with a warm-up iteration count of 40. The model is iterated 300 epochs on ImageNet-1k and ImageNet-100, and 1,000 epochs on CIFAR-10 and CIFAR-100.

The effectiveness of the algorithms is evaluated by freezing the feature encoding network and training only the linear layer. For ImageNet-1k, during the linear evaluation phase, the batch size is 512 with a learning rate of 3E-4, and the self-supervised models trained for 200 epochs are assessed using the top 1 classification metrics.

\begin{table}[htbp]
\centering
\caption{The comparison of the proposed method with ResNet-50 as the backbone under different numbers of pre training iterations.}
\label{table3}
\begin{tabular}{lccccc}
\toprule
Method&Batch size & Backbone& 100 ep & 200 ep & 400 ep \\
\midrule
         SimCLR\cite{chen_simple_2020} &4096 &  ResNet-50 &66.5   &68.3 &       69.8         \\
          SwAV\cite{caron_unsupervised_2020} &4096 &ResNet-50 &  66.5 &69.1 &70.7 \\
          MoCo-v2\cite{chen_improved_2020} &256 &  ResNet-50  &67.4     & 69.9     &71.0        \\
         SimSiam\cite{chen_exploring_2020} &256&    ResNet-50   &68.1  &  70.0         &70.8   \\

           Ours&  256&     ResNet-50    &  \textbf{68.3} &   \textbf{70.9}     &\textbf{71.1}     \\

\bottomrule
\end{tabular}
\end{table}

\begin{algorithm}[!ht]
\small
\SetAlgoLined
\clearpage
    \PyComment{Q: Q side blocks} \\
    \PyComment{K: K side blocks} \\
    \PyComment{m: momentum hyperparameter } \\
    \PyComment{t: temperature } \\
    \PyComment{mm: matrix multiplication } \\
     \PyComment{patchify: patch partition} \\
     \PyComment{unpatchify: patch restore} \\
    \clearpage
    \PyCode{for x in loader} \\
    \Indp   
        \PyCode{x1= aug(x)} \\
        \PyCode{x2 = BA(aug(x))}
        
        \PyCode{q1, q2 = Q(x1), Q(x2)} \\
        \PyCode{k1, k2 = K(x1), K(x2)} \\
         \PyComment{symmetrized loss} \\
        \PyCode{loss = ctr(q1, k2) + ctr(q2, k1)} \\
        \PyCode{loss.backward()} \\
         \PyComment{update Q and BA} \\
         \PyCode{update(Q, BA)} \\
         \PyComment{ momentum update K} \\
         \PyCode{K = m*K + (1-m)*Q} \\
    \Indm 
    \clearpage
    \PyComment{Batch Adaptive} \\
    \PyCode{def BA(x):} \\
     \Indp
     \PyComment{(b,c,w,h)->(b,w/p*h/p,p*p*c)}\\
         \PyCode{x\_patch = patchify(x)}  \\ 
         \PyComment{(1,b,w/p*h/p,p*p*c)} \\
         \PyCode{x\_patch = x\_patch.unsqueeze(0)} \\ 
         \PyComment{conv and filter} \\
         \PyCode{x\_f = conv\_embedding(x\_patch)}  \\
         \PyComment{(b,w/p*h/p,p*p*c)} \\
         \PyCode{x\_f = x\_f.squeeze(0)} \\
         \PyComment{patch restore to (b, c, w, h)} \\
         \PyCode{x\_aug = x + unpatchify(x\_f)}  \\
         \PyCode{return x\_aug} \\
     \Indm
     \clearpage
     \PyComment{contrastive loss} \\
     \PyCode{def ctr(q, k):} \\
     \Indp
         \PyCode{logits = mm(q, k.transpose)} \\
         \PyCode{labels = range(N)} \\
         \PyCode{loss = CE\_Loss(logits/t, labels)} \\
         \PyCode{return 2 * t* loss } \\
      \Indm
   \clearpage
\caption{PyTorch pseudo-code of BA-SSL}
\label{algorithm1}
\end{algorithm}

\subsection{Main Results}
\paragraph{The Effectiveness of the Proposed Method.}We use a dual-track self-supervised learning framework containing positive and negative samples by combining the advantageous features of self-supervised learning such as MoCo series, BYOL, and SimCLR. The performance of the proposed method on ImageNet-1k, ImagNet-100, CIFAR-10, and CIFAR-100 can be shown in Table \ref{table1} and Table \ref{table2}. We further reduce the distance between self-supervised learning and supervised learning. As shown in  Table \ref{table2}, we improve the top1 baseline of 52.5\% based on MoCo-v2 by nearly 7 percentage points to 59.41\% top1 accuracy with ResNet-18 as a self-supervised encoder, which significantly closes the baseline gap of supervised learning based on ResNet-18. We also obtained 63.62\%, 92.12\%, and 66.53\% top 1 accuracies for 300 or 1000 epochs in the dataset evaluations of ImageNet-100, CIFAR-10, and CIFAR-100. It should be noted that we used four NVIDIA RTX 3090 (24G) for model training, and due to limited computational resources, we could not verify the effect of a super large batch on the model results, we hope that this part of the study can be verified in the future.
\paragraph{Comparison with Other SSL Frameworks.} To fairly compare different SSL methods, we retrained different self-supervised learning methods using setting the same batch size with epoch. As shown in Table \ref{table2}, on the benchmark test of ImageNet-1k, the method proposed in this paper achieves the leading performance after 300 pre-trained epochs and 200 linear evaluation epochs, which even outperforms those need excessive batch sizes SSL methods. As for ImageNet-100, CIFAR-10 and CIFAR-100, we still achieve competitive test results with full convergence of the model, as shown in Table \ref{table2}. To further illustrate the effectiveness of the proposed method in the setting of small data batches, we compared the metrics for ImagNet-1k under different pretrained epochs when ResNet-50 is backbone in Table \ref{table3}.

\begin{table}[ht]
\centering
\caption{We performed 10 tests of several SOTA solutions and reported the best performance results, with a maximum improvement of 1.25\% on MoCo-v2 for top1 acc.}
\label{table4}

\begin{tabular}{l|l|l|l}
\toprule
Method & Top1  &  Batchsize & Epoch \\
\midrule
  MoCo-v2&  66.29   &    256  &400 \\
 BYOL& 67.95     & 256  & 400 \\
   SimCLR&  63.34 &   256   &400  \\
    SimSiam&  66.25 &   256  &400 \\
    \midrule
    MoCo-v2 + BA&  67.54 \textcolor{red}{($\uparrow$ 1.25)} &256   &400 \\
        BYOL + BA&  68.76 \textcolor{red}{($\uparrow$ 0.81)} & 256  &400 \\
            SimCLR + BA&64.02 \textcolor{red}{($\uparrow$ 0.68)}   &256    &400 \\
                SimSiam + BA&66.97 \textcolor{red}{($\uparrow$ 0.72)}    &   256 & 400\\
\bottomrule
\end{tabular}
\end{table}

\subsection{Plug and Play}
Since the method presented in this paper is proposed to be used for batch fusion in the preloading process of data, it can be used as a plug-and-play method for enhancing other self-supervised learning models to achieve higher levels of test accuracy. As shown in Table \ref{table4}, we inserted the adaptive batch fusion technique proposed in this paper into several classical two-track self-supervised learning models and achieved up to 1.25\% improvement on ImageNet-100 with the default number of batches.

\subsection{The Impact of Embedding Layer Number}
By default, we used only a single Embedding Layer with the structure shown in Figure \ref{figure3} but it can be nested to increase and continuously enhance the initial representation of the data used for self-supervised learning. Therefore, we explored the effect of the number of Embedding layers on the effectiveness of self-supervised learning. As shown in Table \ref{table5}, in ImagNet-1k, we trained and tested the model with different numbers of $Embedding Layers$ using MoCo-v2 as the self-supervised framework. The model performance drops when the number is set to 2, and the model effect picks up when it is 3, but it is never as good as the default setting of Embedding Layer with a single structure. We speculate that this may be due to excessive information feedforward causing the feature encoding network, which is used for training and testing, to be unable to robustly utilise the effective information.

\begin{table}[ht]
\centering
\caption{The BA-SSL get the \textbf{best} representation performance when the number of $Embedding Layer$ is 1. Over-adding this block may cause information leakage that Impairment of data presentation.}
\label{table5}

\begin{tabular}{l|l|c|c}
\toprule
 Method &Backbone & Layer & Top1 \\
\midrule
Bare(MoCo-v2) &ResNet-18 &0 &52.50 \\
 &ResNet-18&1 &\textbf{53.35} \\
 &ResNet-18 &2 &52.62 \\
 &ResNet-18 &3 &53.17 \\
\bottomrule
\end{tabular}
\end{table}

\SetAlgoSkip{smallskip}

\section{Conclusion, Discussion, Limitation and Future Work}
In this paper, we start from the two basic tasks in self-supervised contrastive learning and point out that both the pretexting task and the data batch size are important factors affecting self-supervised learning. We propose a batch fusion adaptive self-supervised learning method based on batch fusion, BA-SSL, that takes the above factors into account effectively. After extensive and fair tests and comparisons, this paper demonstrates the effectiveness of the proposed method in self-supervised contrastive learning.  Meanwhile, the BA-SSL method proposed in this paper achieves leading performance under the same conditions by comparative validation of self-supervised learning in small batches. In addition, it can be applied as a plug-and-play batch fusion technique to existing SOTA methods and bring performance improvement to them. The BA-SSL method proposed in this paper is expected to take advantage of different self-supervised learning paradigms with smaller sample batches and bring new perspectives to the data-driven self-supervised learning community.

\paragraph{Challenges and Solutions in SSL}
As presented earlier, self-supervised learning is still currently affected by both influences from batch and agent task design, and how to play both influences effectively remains an urgent problem in the field of self-supervised learning. Currently, there is a large amount of research work on pretext task design. However, due to the exponentially growing computational space, it is still a promising research direction for building small-batch lightweight training models\cite{lin_establishing_2023}. Especially in the rapidly developing society of interactive artificial intelligence, it is especially important to mine effective feature representations using smaller batches of data. We propose a batch fusion technique in this paper, which requires only simple Patchization and Reconstruction to achieve fusion through multi-level Convs, but due to the limited computational resources, we are not able to verify the significant effect on those very large batches. We hope that this part of the work can inspire later studies. In addition, we noticed that through batch fusion, those cross-modal data (e.g., medical images\cite{behrad_overview_2022} such as CT, US, etc.) seem to be able to be put together and analysed, which would be a possible research direction based on the techniques in this paper. As multimodal technologies are widely used in the development of society, the method proposed in this paper is also expected to provide new research ideas between cross-modal data.

\clearpage  

%
%
\bibliographystyle{splncs04}
\bibliography{eccv2024}

\begin{thebibliography}{10}
\providecommand{\url}[1]{\texttt{#1}}
\providecommand{\urlprefix}{URL }
\providecommand{\doi}[1]{https://doi.org/#1}

\bibitem{bao_beit_2022}
Bao, H., Dong, L., Piao, S., Wei, F.: {BEiT}: {BERT} {Pre}-{Training} of
  {Image} {Transformers} (Sep 2022). \doi{10.48550/arXiv.2106.08254},
  \url{http://arxiv.org/abs/2106.08254}, arXiv:2106.08254 [cs]

\bibitem{bardes_vicreg_2022}
Bardes, A., Ponce, J., LeCun, Y.: {VICReg}:
  {Variance}-{Invariance}-{Covariance} {Regularization} for {Self}-{Supervised}
  {Learning} (Jan 2022). \doi{10.48550/arXiv.2105.04906},
  \url{http://arxiv.org/abs/2105.04906}, arXiv:2105.04906 [cs]

\bibitem{behrad_overview_2022}
Behrad, F., Saniee~Abadeh, M.: An overview of deep learning methods for
  multimodal medical data mining. Expert Systems with Applications
  \textbf{200},  117006 (Aug 2022). \doi{10.1016/j.eswa.2022.117006},
  \url{https://www.sciencedirect.com/science/article/pii/S0957417422004249}

\bibitem{caron_deep_2018}
Caron, M., Bojanowski, P., Joulin, A., Douze, M.: Deep {Clustering} for
  {Unsupervised} {Learning} of {Visual} {Features}. pp. 132--149 (2018),
  \url{https://openaccess.thecvf.com/content_ECCV_2018/html/Mathilde_Caron_Deep_Clustering_for_ECCV_2018_paper.html}

\bibitem{caron_unsupervised_2020}
Caron, M., Misra, I., Mairal, J., Goyal, P., Bojanowski, P., Joulin, A.:
  Unsupervised {Learning} of {Visual} {Features} by {Contrasting} {Cluster}
  {Assignments}. In: Advances in {Neural} {Information} {Processing} {Systems}.
  vol.~33, pp. 9912--9924. Curran Associates, Inc. (2020),
  \url{https://proceedings.neurips.cc/paper/2020/hash/70feb62b69f16e0238f741fab228fec2-Abstract.html}

\bibitem{caron_emerging_2021}
Caron, M., Touvron, H., Misra, I., J{\'e}gou, H., Mairal, J., Bojanowski, P.,
  Joulin, A.: Emerging {Properties} in {Self}-{Supervised} {Vision}
  {Transformers}. pp. 9650--9660 (2021),
  \url{https://openaccess.thecvf.com/content/ICCV2021/html/Caron_Emerging_Properties_in_Self-Supervised_Vision_Transformers_ICCV_2021_paper.html}

\bibitem{chen_simple_2020}
Chen, T., Kornblith, S., Norouzi, M., Hinton, G.: A {Simple} {Framework} for
  {Contrastive} {Learning} of {Visual} {Representations} (Jun 2020).
  \doi{10.48550/arXiv.2002.05709}, \url{http://arxiv.org/abs/2002.05709},
  arXiv:2002.05709 [cs, stat]

\bibitem{chen_big_2020}
Chen, T., Kornblith, S., Swersky, K., Norouzi, M., Hinton, G.: Big
  {Self}-{Supervised} {Models} are {Strong} {Semi}-{Supervised} {Learners} (Oct
  2020). \doi{10.48550/arXiv.2006.10029},
  \url{http://arxiv.org/abs/2006.10029}, arXiv:2006.10029 [cs, stat]

\bibitem{chen_context_2022}
Chen, X., Ding, M., Wang, X., Xin, Y., Mo, S., Wang, Y., Han, S., Luo, P.,
  Zeng, G., Wang, J.: Context {Autoencoder} for {Self}-{Supervised}
  {Representation} {Learning} (May 2022). \doi{10.48550/arXiv.2202.03026},
  \url{http://arxiv.org/abs/2202.03026}, arXiv:2202.03026 [cs]

\bibitem{chen_improved_2020}
Chen, X., Fan, H., Girshick, R., He, K.: Improved {Baselines} with {Momentum}
  {Contrastive} {Learning} (Mar 2020). \doi{10.48550/arXiv.2003.04297},
  \url{http://arxiv.org/abs/2003.04297}, arXiv:2003.04297 [cs]

\bibitem{chen_exploring_2020}
Chen, X., He, K.: Exploring {Simple} {Siamese} {Representation} {Learning} (Nov
  2020). \doi{10.48550/arXiv.2011.10566},
  \url{http://arxiv.org/abs/2011.10566}, arXiv:2011.10566 [cs]

\bibitem{chen_empirical_2021}
Chen, X., Xie, S., He, K.: An {Empirical} {Study} of {Training}
  {Self}-{Supervised} {Vision} {Transformers} (Aug 2021).
  \doi{10.48550/arXiv.2104.02057}, \url{http://arxiv.org/abs/2104.02057},
  arXiv:2104.02057 [cs]

\bibitem{5206848}
Deng, J., Dong, W., Socher, R., Li, L.J., Li, K., Fei-Fei, L.: Imagenet: A
  large-scale hierarchical image database. In: 2009 IEEE Conference on Computer
  Vision and Pattern Recognition. pp. 248--255 (2009).
  \doi{10.1109/CVPR.2009.5206848}

\bibitem{dosovitskiy2020image}
Dosovitskiy, A., Beyer, L., Kolesnikov, A., Weissenborn, D., Zhai, X.,
  Unterthiner, T., Dehghani, M., Minderer, M., Heigold, G., Gelly, S., et~al.:
  An image is worth 16x16 words: Transformers for image recognition at scale.
  arXiv preprint arXiv:2010.11929  (2020)

\bibitem{ermolov_whitening_2021}
Ermolov, A., Siarohin, A., Sangineto, E., Sebe, N.: Whitening for
  {Self}-{Supervised} {Representation} {Learning}. In: Proceedings of the 38th
  {International} {Conference} on {Machine} {Learning}. pp. 3015--3024. PMLR
  (Jul 2021), \url{https://proceedings.mlr.press/v139/ermolov21a.html}, iSSN:
  2640-3498

\bibitem{garrido_duality_2022}
Garrido, Q., Chen, Y., Bardes, A., Najman, L., Lecun, Y.: On the duality
  between contrastive and non-contrastive self-supervised learning (Oct 2022).
  \doi{10.48550/arXiv.2206.02574}, \url{http://arxiv.org/abs/2206.02574},
  arXiv:2206.02574 [cs]

\bibitem{grill_bootstrap_2020}
Grill, J.B., Strub, F., Altch{\'e}, F., Tallec, C., Richemond, P.H.,
  Buchatskaya, E., Doersch, C., Pires, B.A., Guo, Z.D., Azar, M.G., Piot, B.,
  Kavukcuoglu, K., Munos, R., Valko, M.: Bootstrap your own latent: {A} new
  approach to self-supervised {Learning} (Sep 2020).
  \doi{10.48550/arXiv.2006.07733}, \url{http://arxiv.org/abs/2006.07733},
  arXiv:2006.07733 [cs, stat]

\bibitem{he_masked_2021}
He, K., Chen, X., Xie, S., Li, Y., Doll{\'a}r, P., Girshick, R.: Masked
  {Autoencoders} {Are} {Scalable} {Vision} {Learners} (Dec 2021).
  \doi{10.48550/arXiv.2111.06377}, \url{http://arxiv.org/abs/2111.06377},
  arXiv:2111.06377 [cs]

\bibitem{he_momentum_2020}
He, K., Fan, H., Wu, Y., Xie, S., Girshick, R.: Momentum {Contrast} for
  {Unsupervised} {Visual} {Representation} {Learning}. Tech. Rep.
  arXiv:1911.05722, arXiv (Mar 2020), \url{http://arxiv.org/abs/1911.05722},
  arXiv:1911.05722 [cs] type: article

\bibitem{huang_revisiting_2022}
Huang, J., Kong, X., Zhang, X.: Revisiting the {Critical} {Factors} of
  {Augmentation}-{Invariant} {Representation} {Learning}. In: Avidan, S.,
  Brostow, G., Ciss{\'e}, M., Farinella, G.M., Hassner, T. (eds.) Computer
  {Vision} -- {ECCV} 2022. pp. 42--58. Lecture {Notes} in {Computer} {Science},
  Springer Nature Switzerland, Cham (2022). \doi{10.1007/978-3-031-19821-2_3}

\bibitem{krizhevsky2009learning}
Krizhevsky, A., Hinton, G., et~al.: Learning multiple layers of features from
  tiny images  (2009)

\bibitem{li2023minent}
Li, S., Liu, F., Hao, Z., Jiao, L., Liu, X., Guo, Y.: Minent: Minimum entropy
  for self-supervised representation learning. Pattern Recognition
  \textbf{138},  109364 (2023)

\bibitem{li2022self}
Li, S., Liu, F., Jiao, L., Chen, P., Li, L.: Self-supervised self-organizing
  clustering network: a novel unsupervised representation learning method. IEEE
  Transactions on Neural Networks and Learning Systems  (2022)

\bibitem{lin_establishing_2023}
Lin, W., Ding, Y., Cao, Z., Zheng, H.T.: Establishing a {Stronger} {Baseline}
  for {Lightweight} {Contrastive} {Models}. In: 2023 {IEEE} {International}
  {Conference} on {Multimedia} and {Expo} ({ICME}). pp. 1062--1067 (2023).
  \doi{10.1109/ICME55011.2023.00186}

\bibitem{mo_multi-level_2023}
Mo, S., Sun, Z., Li, C.: Multi-{Level} {Contrastive} {Learning} for
  {Self}-{Supervised} {Vision} {Transformers}. pp. 2778--2787 (2023),
  \url{https://openaccess.thecvf.com/content/WACV2023/html/Mo_Multi-Level_Contrastive_Learning_for_Self-Supervised_Vision_Transformers_WACV_2023_paper.html}

\bibitem{nguyen_dimcl_2023}
Nguyen, T., Pham, T.X., Zhang, C., Luu, T.M., Vu, T., Yoo, C.D.: {DimCL}:
  {Dimensional} {Contrastive} {Learning} for {Improving} {Self}-{Supervised}
  {Learning}. IEEE Access  \textbf{11},  21534--21545 (2023).
  \doi{10.1109/ACCESS.2023.3236087}

\bibitem{oinar_expectation-maximization_2023}
Oinar, C., M.~Le, B., Woo, S.S.: Expectation-{Maximization} via
  {Pretext}-{Invariant} {Representations}. IEEE Access  \textbf{11},
  65266--65276 (2023). \doi{10.1109/ACCESS.2023.3289589},
  \url{https://ieeexplore.ieee.org/abstract/document/10163821}, conference
  Name: IEEE Access

\bibitem{oord2019representation}
van~den Oord, A., Li, Y., Vinyals, O.: Representation learning with contrastive
  predictive coding (2019)

\bibitem{oquab_dinov2_2023}
Oquab, M., Darcet, T., Moutakanni, T., Vo, H., Szafraniec, M., Khalidov, V.,
  Fernandez, P., Haziza, D., Massa, F., El-Nouby, A., Assran, M., Ballas, N.,
  Galuba, W., Howes, R., Huang, P.Y., Li, S.W., Misra, I., Rabbat, M., Sharma,
  V., Synnaeve, G., Xu, H., Jegou, H., Mairal, J., Labatut, P., Joulin, A.,
  Bojanowski, P.: {DINOv2}: {Learning} {Robust} {Visual} {Features} without
  {Supervision} (Apr 2023). \doi{10.48550/arXiv.2304.07193},
  \url{http://arxiv.org/abs/2304.07193}, arXiv:2304.07193 [cs]

\bibitem{ozsoy_self-supervised_2022}
Ozsoy, S., Hamdan, S., Arik, S., Yuret, D., Erdogan, A.: Self-{Supervised}
  {Learning} with an {Information} {Maximization} {Criterion}. Advances in
  Neural Information Processing Systems  \textbf{35},  35240--35253 (Dec 2022),
  \url{https://proceedings.neurips.cc/paper_files/paper/2022/hash/e4cd50120b6d7e8daff1749d6bbaa889-Abstract-Conference.html}

\bibitem{10.1007/978-3-031-19821-2_25}
Rizve, M.N., Kardan, N., Shah, M.: Towards realistic semi-supervised learning.
  In: Avidan, S., Brostow, G., Ciss{\'e}, M., Farinella, G.M., Hassner, T.
  (eds.) Computer Vision -- ECCV 2022. pp. 437--455. Springer Nature
  Switzerland, Cham (2022)

\bibitem{tian_contrastive_2020}
Tian, Y., Krishnan, D., Isola, P.: Contrastive {Multiview} {Coding}. In:
  Vedaldi, A., Bischof, H., Brox, T., Frahm, J.M. (eds.) Computer {Vision} --
  {ECCV} 2020. pp. 776--794. Lecture {Notes} in {Computer} {Science}, Springer
  International Publishing, Cham (2020). \doi{10.1007/978-3-030-58621-8_45}

\bibitem{tian_what_2020}
Tian, Y., Sun, C., Poole, B., Krishnan, D., Schmid, C., Isola, P.: What {Makes}
  for {Good} {Views} for {Contrastive} {Learning}? In: Advances in {Neural}
  {Information} {Processing} {Systems}. vol.~33, pp. 6827--6839. Curran
  Associates, Inc. (2020),
  \url{https://proceedings.neurips.cc/paper/2020/hash/4c2e5eaae9152079b9e95845750bb9ab-Abstract.html}

\bibitem{tong_emp-ssl_2023}
Tong, S., Chen, Y., Ma, Y., Lecun, Y.: {EMP}-{SSL}: {Towards}
  {Self}-{Supervised} {Learning} in {One} {Training} {Epoch} (Apr 2023).
  \doi{10.48550/arXiv.2304.03977}, \url{http://arxiv.org/abs/2304.03977},
  arXiv:2304.03977 [cs]

\bibitem{wei_masked_2022}
Wei, C., Fan, H., Xie, S., Wu, C.Y., Yuille, A., Feichtenhofer, C.: Masked
  {Feature} {Prediction} for {Self}-{Supervised} {Visual} {Pre}-{Training}. pp.
  14668--14678 (2022),
  \url{https://openaccess.thecvf.com/content/CVPR2022/html/Wei_Masked_Feature_Prediction_for_Self-Supervised_Visual_Pre-Training_CVPR_2022_paper.html}

\bibitem{wu_extreme_2023}
Wu, Z., Lai, Z., Sun, X., Lin, S.: Extreme {Masking} for {Learning} {Instance}
  and {Distributed} {Visual} {Representations} (Mar 2023).
  \doi{10.48550/arXiv.2206.04667}, \url{http://arxiv.org/abs/2206.04667},
  arXiv:2206.04667 [cs]

\bibitem{xie_simmim_2022}
Xie, Z., Zhang, Z., Cao, Y., Lin, Y., Bao, J., Yao, Z., Dai, Q., Hu, H.:
  {SimMIM}: {A} {Simple} {Framework} for {Masked} {Image} {Modeling}. pp.
  9653--9663 (2022),
  \url{https://openaccess.thecvf.com/content/CVPR2022/html/Xie_SimMIM_A_Simple_Framework_for_Masked_Image_Modeling_CVPR_2022_paper.html}

\bibitem{zbontar_barlow_2021}
Zbontar, J., Jing, L., Misra, I., LeCun, Y., Deny, S.: Barlow {Twins}:
  {Self}-{Supervised} {Learning} via {Redundancy} {Reduction}. In: Proceedings
  of the 38th {International} {Conference} on {Machine} {Learning}. pp.
  12310--12320. PMLR (Jul 2021),
  \url{https://proceedings.mlr.press/v139/zbontar21a.html}, iSSN: 2640-3498

\bibitem{zhou_ibot_2022}
Zhou, J., Wei, C., Wang, H., Shen, W., Xie, C., Yuille, A., Kong, T.: {iBOT}:
  {Image} {BERT} {Pre}-{Training} with {Online} {Tokenizer} (Jan 2022).
  \doi{10.48550/arXiv.2111.07832}, \url{http://arxiv.org/abs/2111.07832},
  arXiv:2111.07832 [cs]

\bibitem{zhou_mimco_2022}
Zhou, Q., Yu, C., Luo, H., Wang, Z., Li, H.: {MimCo}: {Masked} {Image}
  {Modeling} {Pre}-training with {Contrastive} {Teacher}. In: Proceedings of
  the 30th {ACM} {International} {Conference} on {Multimedia}. pp. 4487--4495.
  {MM} '22, Association for Computing Machinery, New York, NY, USA (Oct 2022).
  \doi{10.1145/3503161.3548173}, \url{https://doi.org/10.1145/3503161.3548173}

\end{thebibliography}
\end{document}